\documentclass{article}

    \usepackage[preprint]{neurips_2020}

\usepackage[utf8]{inputenc} 
\usepackage[T1]{fontenc}    
\usepackage{hyperref}       
\usepackage{url}            
\usepackage{booktabs}       
\usepackage{amsfonts}       
\usepackage{nicefrac}       
\usepackage{microtype}      

\usepackage[pdftex]{graphicx}
\usepackage[toc,page]{appendix}
\usepackage{listings}

\usepackage{xcolor}

\definecolor{codegreen}{rgb}{0,0.6,0}
\definecolor{codegray}{rgb}{0.5,0.5,0.5}
\definecolor{codepurple}{rgb}{0.58,0,0.82}
\definecolor{backcolour}{rgb}{0.95,0.95,0.92}

\lstdefinestyle{mystyle}{
    backgroundcolor=\color{backcolour},   
    commentstyle=\color{codegreen},
    keywordstyle=\color{magenta},
    numberstyle=\tiny\color{codegray},
    stringstyle=\color{codepurple},
    basicstyle=\ttfamily\footnotesize,
    breakatwhitespace=false,         
    breaklines=true,                 
    captionpos=b,                    
    keepspaces=true,                 
    numbers=right,                    
    numbersep=5pt,                  
    showspaces=false,                
    showstringspaces=false,
    showtabs=false,                  
    tabsize=2
}

\title{Add a SideNet to your MainNet}

\author{%
  Adrien Morisot \\
  Independent Researcher\\
  \texttt{adrien.morisot@gmail.com} \\
}

\begin{document}

\maketitle

\begin{abstract}
  As the performance and popularity of deep neural networks has increased, so too has their computational cost. There are many effective techniques for reducing a network’s computational footprint--quantisation, pruning, knowledge distillation--, but these lead to models whose computational cost is the same regardless of their input. Our human reaction times vary with the complexity of the tasks we perform: easier tasks--e.g. telling apart dogs from boat--are executed much faster than harder ones--e.g. telling apart two similar-looking breeds of dogs. Driven by this observation, we develop a method for adaptive network complexity by attaching a small classification layer, which we call SideNet, to a large pretrained network, which we call MainNet. Given an input, the SideNet returns a classification if its confidence level, obtained via softmax, surpasses a user-determined threshold, and only passes it along to the large MainNet for further processing if its confidence is too low. This allows us to flexibly trade off the network’s performance with its computational cost. Experimental results show that simple single hidden layer perceptron SideNets added onto pretrained ResNet and BERT MainNets allow for substantial decreases in compute with minimal drops in performance on image and text classification tasks. We also highlight three other desirable properties of our method, namely that the classifications obtained by SideNets are calibrated, complementary to other compute-reduction techniques, and that they enable the easy exploration of compute-accuracy space. 
\end{abstract}

\section{Introduction}

In recent years, neural networks have increased dramatically in size: \cite{amodei2018ai} estimate a 300,000x growth in compute since 2012, with a doubling period of 3.4 months. Since the ``bitter lesson'' \citep{sutton2019bitter} of machine learning seems for now to be true, and performance on machine learning tasks appears to scale linearly with model size and amount of training data \citep{hestness2017deep}, this trend is unlikely to decelerate anytime soon. Indeed, at the time of writing, OpenAI has just published GPT-3, a language modelling neural network with 175 Billion parameters \citep{brown2020language}. Since neural networks are increasingly used in industry to power various large-scale applications--from voice recognition (Google's Assistant and Apple's Siri are powered by neural networks \citep{team2017hey, He_2019}) to image processing and natural language understanding--lowering the computational cost of these models at inference time is an increasingly pressing problem. There are ways of reducing the computational footprints of neural networks. Neural network pruning removes less important connections between neurons \citep{castellano1997iterative}. Quantisation reduces the number of bits taken up by each of the network's parameters \citep{han2015deep}. Knowledge distillation uses a larger network to train a smaller network on the large one's outputs \citep{hinton2015distilling}. 

These methods, although powerful, still lead to networks spending the same amount of compute on each input, regardless of the complexity of the input. Yet humans take different amounts of time to solve different tasks based on the complexity of the tasks (it is easier to quickly distinguish between a bear and a boat than it is to quickly distinguish between an Alaskan Malamute and a Siberian Husky). This observation gave rise to the field of conditional computation. Conditional computation techniques involve incorporating mechanisms within neural networks that allow the networks to reduce inference time compute costs by not passing the input through the entire graph, but only a small sub-part of it. This lets the network spend less compute time on easier inputs/tasks, and has the added benefit of allowing the network to be sensitive to computational budgets (if the budget is high, the network can afford to use more compute).

Existing implementations of conditional computation are generally complicated to engineer, and consequently are not used much in industry \citep{bapna2020controlling}. 

To solve this, we propose the simplest model of conditional computation: attaching a single hidden layer perceptron, which we call SideNet, to an intermediate representation of a pretrained network, which we call MainNet. Unlike most existing conditional computation methods, the SideNet is straightforward to train, and attaching a SideNet to a MainNet is easy to engineer. 

We also make three noteworthy observations: (i) When attached to the early intermediate representations of ResNets, the classification confidences of SideNets are calibrated, whereas the classification confidences of their ResNets are not. (ii) SideNet-based compute reduction can be complementary to knowledge-distillation and pruning: applying SideNets to DistilBERT \citep{sanh2019distilbert}, a heavily compressed transformer model, still yielded noticeable performance savings ($\approx 30\%$) for a small drop in test accuracy ($\approx 0.5\%$). (iii) SideNets make it easy to explore compute-accuracy space, by making it continuous rather than discrete.

\section{Related Work}
\label{relwork}

\subsection{Similar architectures}

 \cite{tinytobig} first run an image through a small convolutional neural network to ascertain whether or not it can be classified with high confidence. If it cannot, they send the image to a larger network, and use that classification as the final one. \cite{bolukbasi2017adaptive} build on this. They first run an image through a small AlexNet classifier \citep{alexnet}, and a regression model determines the confidence level of the classification. If it is high, the classification is returned; otherwise, the image is sent through a GoogLeNet classifier \citep{googlenet}, where the same regression is applied. If the confidence is still too low, it is sent through a ResNet \citep{he2016deep}, where a final classification is returned. Our method differs from these because in ours less computation is wasted: if the SideNet's confidence in its prediction is not high enough to return a classification, then the intermediary representation it used will continue flowing along the MainNet, and will not have to be recomputed from scratch.

\cite{leroux2017cascading} is the paper most resembling ours: they run an image through a main backbone network, along with multiple small classification networks along the backbone's side that interrupt the flow of the image through the main model if their confidence is high enough. They demonstrated that their method provided significant energy savings on a Raspberry Pi computer. \cite{zhang2019scan} build on this, by using attention mechanisms with their side classification networks, and training them with knowledge-distillation and a genetic algorithm. Our method differs from these because it only uses one SideNet, which makes training substantially easier (training a network with multiple heads requires properly weighting the losses of each head, which is challenging).

There are a variety of other architectures involving conditional computation: \cite{bengio2015conditional} use reinforcement learning to learn a policy that directs an input only through discrete parts of a network, rather than the whole network. However, backpropagating through discrete random variables is inefficient and slow. \cite{bapna2020controlling} introduce a method to turn these discrete random variables continuous, to increase the rate of learning, and use it to train control networks, networks that control the amount of compute used at inference.

\subsection{Intermediate representations}

There is a rich literature studying neural networks' intermediate representations. 
\cite{alexnet} find that early layers of convolutional neural networks mostly pick out simple textures and lines. This suggests that if an image is texturally simple or distinctive, it should be able to be classified in early parts of the network, rather than at the very end. \cite{leroux2017cascading} argue that this holds: their model was confident in its predictions when the input was fairly straightforward, and passed it off to the deeper model when it was more visually complex (e.g. the digit 1 is less complex than the italicised digit \textit{1}, and was classified earlier in the network).

Similarly, in natural language processing, \cite{clark2019does} find that early layers of BERT (a large transformer architecture by \cite{devlin2018bert}) attend to broad features of an input, as opposed to later layers that tend to focus on a certain particular aspect of an input, and \cite{alex2019emergent} find that <CLS> tokens are heavily overparameterised, and can be  shrunk substantially without affecting performance. 

\section{Method}

\subsection{Framework}
A neural network $M$, at a high level, is a function approximator. It maps inputs $x$ to outputs $y$: $M(x) = y$. Supervised learning involves training the parameters of $M$ to best fit the training data $(x, \tilde{y})$. We can decompose this mapping $M$ into sub-components, and view it as a composition of transformations $M_1$, $M_2$, ..., $M_n$ of the input $x$ into intermediate representations $x_1$, $x_2$, ..., $x_m$. 

For simple architectures, like VGG \citep{simonyan2014very}, the compositions can be written simply, as below: 

$x_1=M_1(x)$, 

$x_2=M_2(x_1)=M_2\circ M_1(x)$, 

...

$y=x_n=M_n\circ M_{n-1}\dots\circ M_2\circ M_1(x)$,

where the $M_i$ are convolutional layers, max-pooling layers, fully connected layers, and non-linear layers. 

More complicated architectures are more involved to formalise, but can nonetheless still be made to fit this framework of intermediate representations. For example, if a layer of the network involves a skip connection from layer $i$ to layer $k$, then we can write $x_k$ and $M_k$ as: 

$x_k=M_k(x_{k-1}, x_i)=x_{k-1}+x_i$. 

\subsection{Architecture}

\begin{figure}

  \centering
  \includegraphics[width=0.90\linewidth]{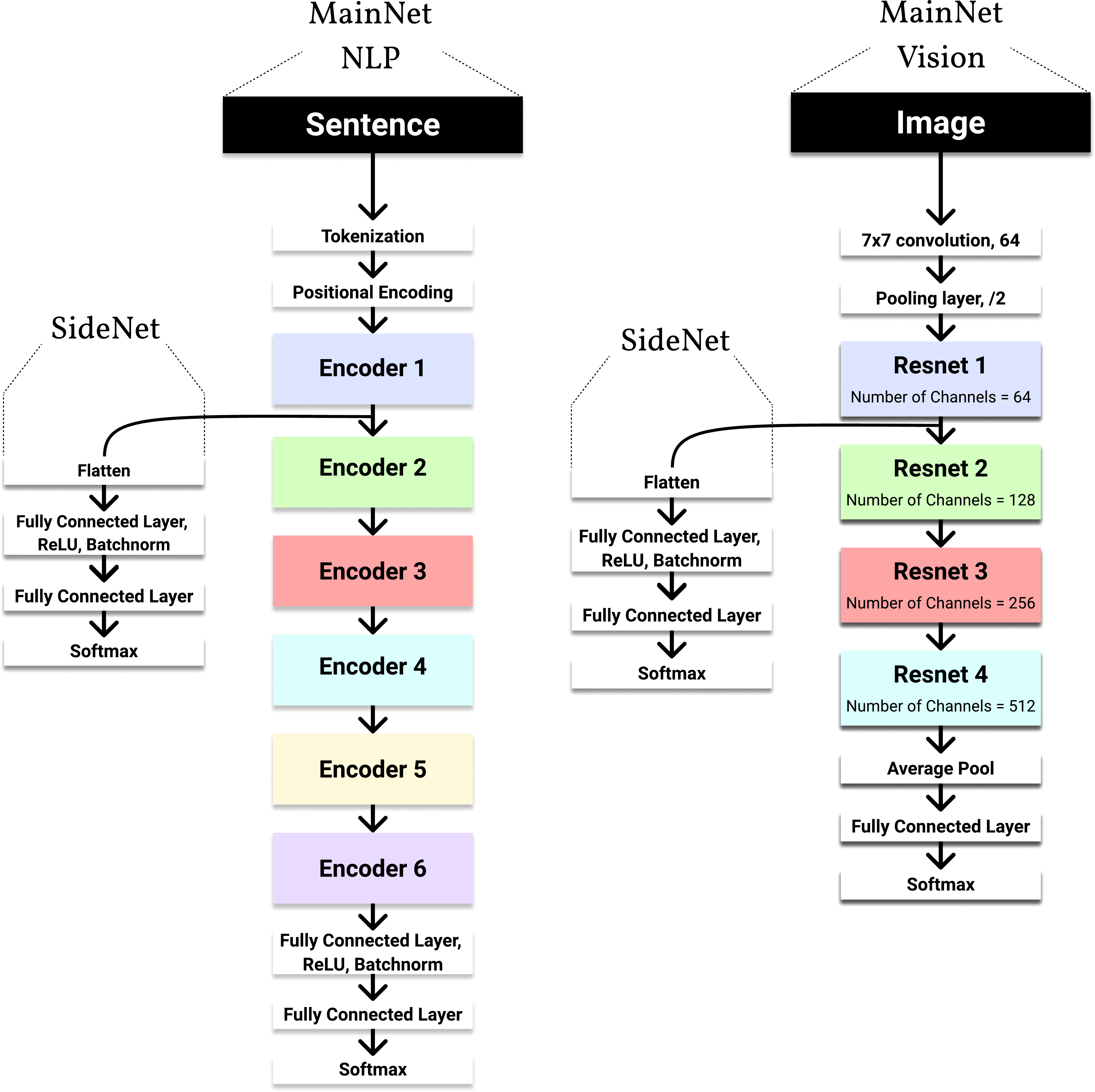}
  \caption{SideNets can be attached to a wide variety of networks. Here we visualise two networks equipped with SideNets. \textit{Left:} A DistilBERT transformer. Its MainNet is untouched, but a SideNet is added between encoders 1 and 2. \textit{Right:} A ResNet. Its MainNet is untouched, but a SideNet identical to the one for DistilBERT is added after its first block.}
  \label{twonets}
\end{figure}

We call the net $M$ the MainNet. On top of this MainNet backbone, we propose to add a SideNet, a simple task-specific network $S$ which takes as input one of the MainNet's intermediary representations $x_c$, and returns a probability distribution over the classes $y_c=S(x_c)$. In our experiments, we purposefully choose $S$ to be extremely simple: a fully connected layer, a non-linear ReLU layer \citep{nair2010rectified}, a batch normalisation layer \citep{ioffe2015batch}, a final fully connected layer, and a softmax layer (or a sigmoid layer in the case of binary classification). Although the softmax operation is not a true reflection of the model's confidence\footnote{See Appendix A for a fuller discussion.} \citep{gal2016uncertainty}, we find that using it as a proxy for model confidence works well empirically. 

SideNets can be attached to any intermediate representation $x_i$; in \autoref{twonets} we illustrate two possible locations for SideNets on two different architectures: the DistilBERT transformer for natural language processing and the ResNet for computer vision.  

\subsection{Training SideNets}

To train the SideNet quickly, we can freeze the weights of the MainNet, and update the SideNet's weights on the normal training data. The SideNet, by construction, has very few parameters, and the input data only needs to flow through a small fraction of the MainNet to get to the SideNet, so the optimisation is fast and converges quickly. Multiple SideNets $S_1$, ..., $S_p$ with parameters $W_1$, ..., $W_p$ can be trained in parallel at different points along the MainNet, as long as they return separate losses $L_{S1}$, ..., $L_{Sp}$, since by construction $\frac{\partial W_i}{\partial L_j}=0,  \forall i \neq j$, provided that the SideNets remain independent of each other. While training this way is significantly faster, it does come with a significant performance cost (on the order of ~3\% in our experiments), so in performance-critical models, fine-tuning the whole model with the SideNet is preferable. 

To fine-tune the weights of the MainNet alongside those of the SideNet, we can backpropagate over the weighted sum of their losses. If the MainNet's loss is $L_M$ and the SideNet's loss is $L_S$, then we can backpropagate over a loss $L = L_M + \alpha L_S$. In our experiments we always pick $\alpha=1$. To show that SideNets are easy to train, we share our code in Appendix B. To fine-tune the weights of the MainNet alongside those of multiple SideNets, each with losses $L_{S1}$, ..., $L_{Sp}$, the same principle applies. 

\subsection{SideNets at inference time}

To classify an input image x, we run x through the MainNet until we obtain the intermediary representation $x_c$, and then pass $x_c$ through $S$ to obtain a classification $\hat{y}$ and confidence level $\hat{p}$. If the confidence level exceeds a user specified threshold $\theta$, then the classification is returned immediately, without having $x_c$ pass through the rest of the MainNet. If the confidence level is below $\theta$, then $x_c$ is passed back on to the MainNet, where it returns a final classification $y$. 

\section{Classification Experiments}

We perform all experiments on a single NVIDIA RTX 2070 GPU. All experiments use the Adam \citep{kingma2014adam} optimiser, with default parameters. We use an initial learning rate of .0003 and train for 50 epochs in the ResNet experiments; we use an initial learning rate of .000003 and train for 20 epochs in the BERT and DistilBERT experiments. In both cases, we use a learning rate decay of 3 after 5 epochs in which the validation loss doesn't go down. 

For all experiments, our SideNet is a single hidden layer perceptron, with an input size equal to the number of elements in $x_c$ (a flattened version of $x_c$ for images), a hidden layer with 32 units,\footnote{We found that increasing this number did not have much of an effect. More details in Appendix C.} a batchnorm layer, a ReLU layer, and a classification layer (softmax for multi-class classification, sigmoid for binary). All of our experimental details can be found in Appendix D. 

\subsection{CIFAR10}
We assess our method's performance on the CIFAR10 dataset \citep{krizhevsky2009learning}, a dataset of 60,000 colour images, 32x32 pixels, with 10 classes of 6,000 elements each. Our train/validate/test split is 50,000/5,000/5,000. We apply standard CIFAR10 data augmentation techniques: normalisation, random cropping with padding 4, and horizontal flips.   

We use ResNet18, 34, 50, and 152 (with weights pretrained on ImageNet) as the core architecture of the MainNet. Since they were pretrained on ImageNet, which has 1,000 classes, we replace their final fully connected layer with a fully connected layer with the same architecture as a SideNet, described above. We attach the SideNet to the output of the Resnet 1 block illustrated in \autoref{twonets}. The SideNet is fine-tuned with the last layer of the MainNet. 

\begin{figure}
  \centering
  \includegraphics[width=0.85\linewidth]{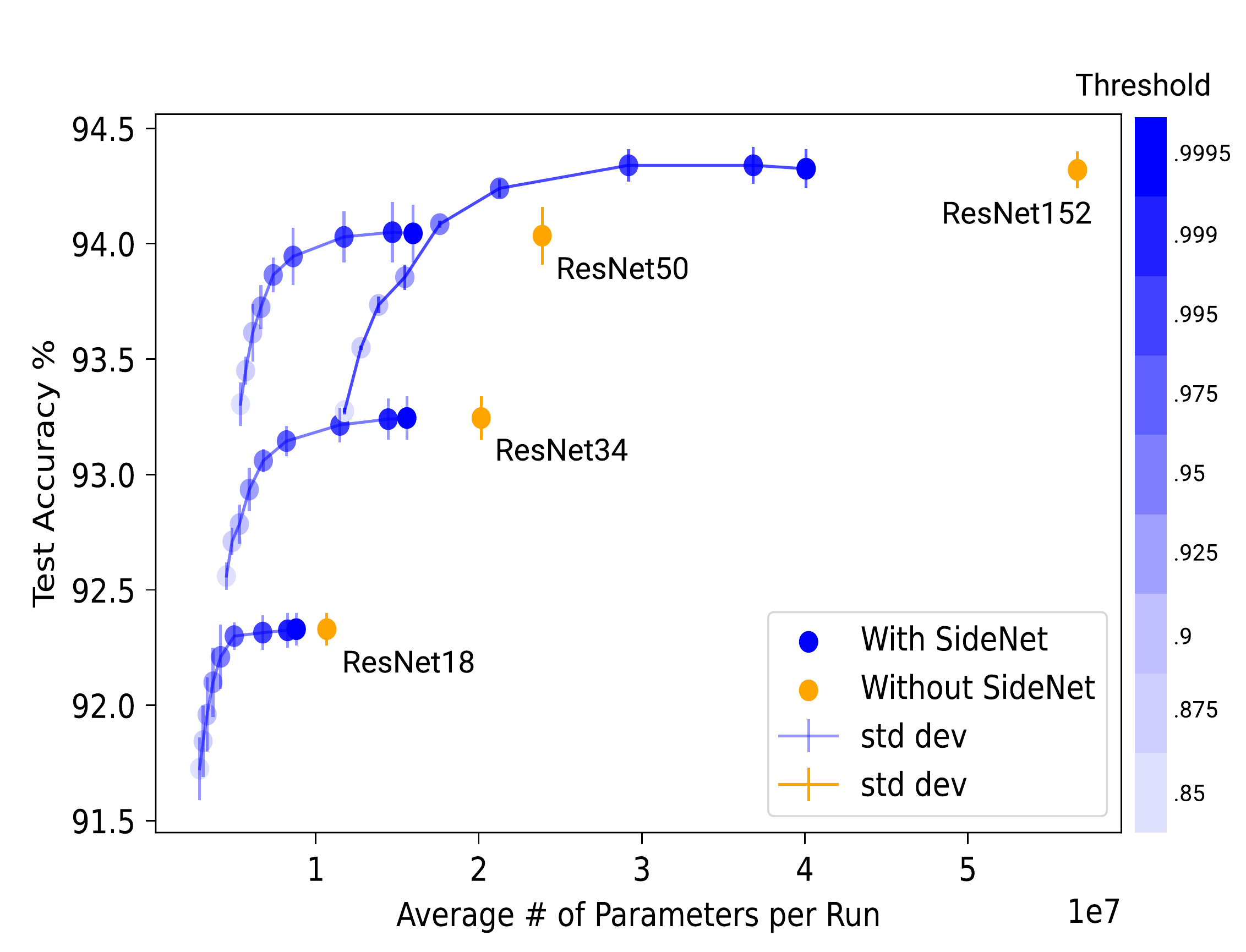}
  \caption{Plot of test accuracy with respect to average number of parameters per run, for different thresholds $\theta$, for different depths of ResNets, with and without SideNets. The results are averaged over 5 runs, with error bars indicating standard deviations.}
  \label{cifartrail}
\end{figure}

We evaluate our method on the test set with different thresholds $\theta$ by plotting the model's accuracy with respect to the amount of compute used. 
We use the average number of parameters used for a single input as a proxy for the amount of compute used (since this number stays fixed, whereas the average number of floating point operations would vary based on the size of the input).
 The results are plotted in \autoref{cifartrail}. We find that architectures using SideNets can use significantly less compute than architectures without SideNets, and still maintain the same accuracy. We also note that adding a SideNet makes it easy and cheap to explore the space of models with different compute and accuracy levels: simply adjust the threshold $\theta$. In order to explore this same compute-accuracy space with knowledge-distillation or pruning, we would have to repeatedly do so from scratch. 

We also test to see if our SideNets are calibrated. A classification model is calibrated when the probability $p$ it assigns to an input $x$ belonging to a certain class is equal to the actual probability of the model classifying it correctly. For example, if a weather model predicts every day for 100 days that it will be sunny with 75\% certainty, and at the end of the 100 days there were indeed 75 sunny days, then that model is calibrated. More formally, given an input $x$ whose true classification label is $y$, if a model $M$ assigns to $x$ a classification of $\hat{y}$ with confidence $\hat{p}$, then $M$ is calibrated iff $\mathbb{P}(\hat{y}=y | \hat{p}=p)=p, \forall p \in [0, 1]$. Calibration is a useful property for a model to have, since it ``knows what it doesn't know''. We quantify calibration using the expected calibration error (ECE). We first bin our predictions into 8 equally spaced classification confidence bins, consisting of $n_i$ predictions each: confidences between 0.2 and 0.3 go into bin 1, ..., confidences between 0.9 and 1 go into bin 8 (there are no bins between 0.1 and 0.2 because in our experiments both SideNets and MainNets always have confidence above 0.2). The ECE is computed by calculating the average distance between confidence and accuracy for each bin: ECE=$\sum_i \frac{n_i}{n} | acc(i) - conf(i) |$, $i=1, \dots, 8$. 

\begin{figure}

  \centering
  \includegraphics[width=0.75\linewidth]{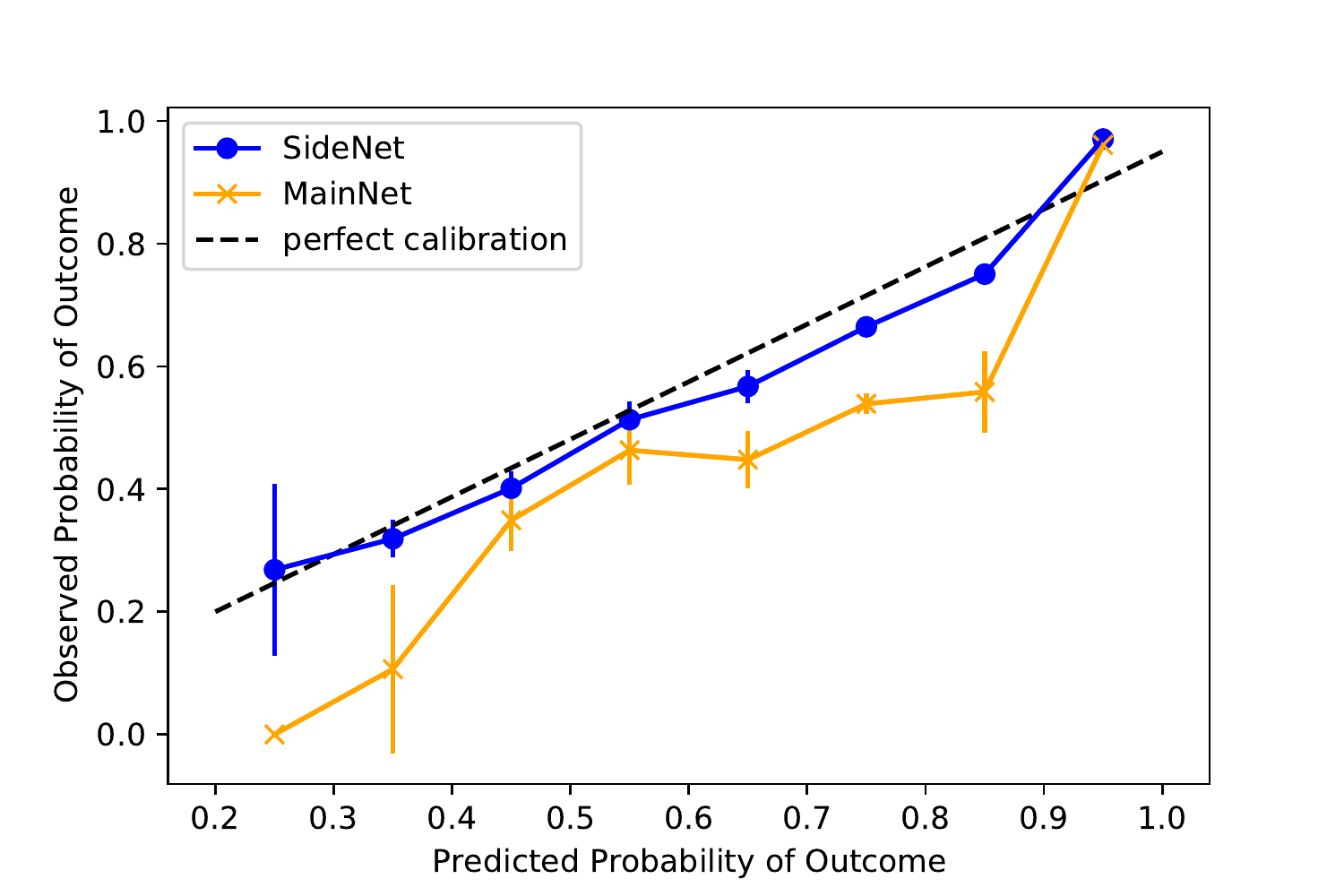}
  \caption{Calibration plot of a ResNet50's SideNet and MainNet, averaged over 5 runs, with standard deviations. The SideNet's classifications are significantly closer to perfect calibration than those of the MainNet. The results were obtained on the test set.}
  \label{calibrn50}
\end{figure}

\cite{guo2017calibration} find that deep convolutional neural networks are not calibrated. We reproduce their results, and find that our MainNet classifications are not calibrated, with high ECE scores. However, the classifications of our SideNets are well calibrated. \autoref{cifar-table} details the ECE scores for SideNets and MainNets, and \autoref{calibrn50} gives a specific example of how the MainNet is uncalibrated relative to the SideNet. This is helpful to the person setting the confidence threshold $\theta$: it means that if they set $\theta=0.85$, then the SideNet will have a minimum accuracy of 85\%.

\begin{table}
  \caption{ECE scores for different SideNets and MainNets, evaluated on the test set. The lower the ECE, the more calibrated the model. The values are averaged over 5 runs, and include standard deviations.}
  \label{cifar-table}
  \centering
  \begin{tabular}{ l | c c }
    \toprule
    Model & SideNet ECE & MainNet ECE  \\
    \midrule
    ResNet152 & $\textbf{.30} \pm .06$ & $1.1 \pm .18$  \\
    ResNet50 & $\textbf{.41} \pm .10$ & $.91 \pm .17$  \\
    ResNet34 & $\textbf{.38} \pm .08$ & $1.0 \pm .14$\\
    ResNet18 & $\textbf{.31} \pm .08$ & $1.0 \pm .05$  \\
    \bottomrule
  \end{tabular}
\end{table}

\subsection{SST-2}

\begin{figure}
  \centering
  \includegraphics[width=0.85\linewidth]{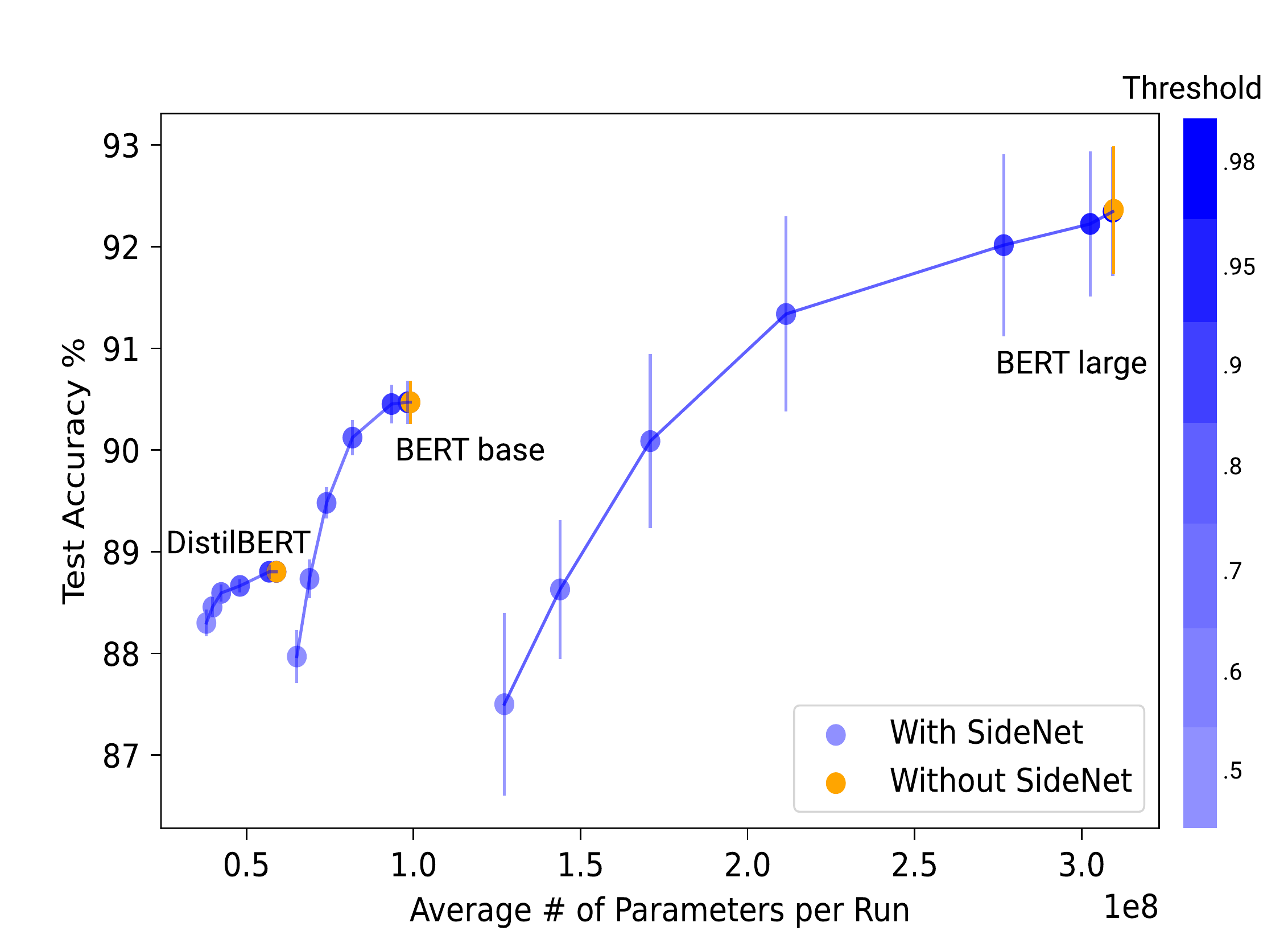}
  \caption{Plot of test accuracy with respect to average number of parameters per run, for different thresholds $\theta$, for different transformer models, with and without SideNets. The results are averaged over 5 runs, with error bars indicating standard deviations.}
  \label{moviereviews}

\end{figure}

To assess our method's performance on natural language processing tasks, we apply it to the SST-2 dataset \citep{socher2013recursive}, a dataset of 9613 movie reviews, labelled as positive or negative.  Our train/validation/test split is 5000/1613/3000. 

We use pretrained DistilBERT \citep{sanh2019distilbert}, BERT-base, and BERT-large \citep{devlin2018bert} models as the core architectures of the MainNet. DistilBERT has 6 encoders with 12 attention heads each, BERT-base has 12 encoders with 12 attention heads each, and BERT-large has 24 encoders with 16 attention heads each. 
For the MainNet's final classification layer, we add a fully connected layer with the same architecture as a SideNet. We attach DistilBERT's SideNet after its first encoder (out of 6), as in \autoref{twonets}; we attach BERT-base's SideNet after its fourth encoder (out of 12); we attach BERT-large's SideNet after its eighth encoder (out of 24).\footnote{For full diagrams of the architectures, see Appendix E.} The SideNet is fine-tuned along with the last layer of the MainNet. 

BERT-base and DistilBERT use 768 dimensional tensors to represent each token, and so the total parameter count overhead of the SideNet is $768 \times 32 + 32\times 1 \approx 25000$, which is $\approx 0.03\%$ of BERT-base's $\approx$100M parameter count, and $\approx 0.04\%$ of DistilBERT's $\approx$60M parameter count. BERT-large uses 1024 dimensional tensors to represent each token, so its overhead is $1024 \times 32+32\times 1 \approx 35000$, which is $\approx 0.01\%$ of BERT-large's $\approx$300M parameter count.

We evaluate on the test set with different thresholds $\theta$, and plot the results in \autoref{moviereviews}, with the same methodology as with \autoref{cifartrail}. We find that adding SideNets allows for substantial decreases in compute, albeit with a greater loss in accuracy than the CIFAR10 example above. However, as above, we note that the addition of SideNets allows for a much easier exploration of compute-accuracy space. If we wanted a model with 200M parameters, rather than the 300M of BERT-large or the 100M of BERT-base, then rather than train that 200M parameter model from scratch, we could easily attach a SideNet to a pretrained BERT-large, and get a model that on average uses 200M parameters per run, with an accuracy above BERT-base, but below BERT-large. 

Furthermore, it is worth highlighting that adding a SideNet to DistilBERT manages to reduce its average parameter use by 30\%, at a cost of ~0.5\% test accuracy, despite it already being a  version of BERT-base that was compressed using extensive model pruning and knowledge distillation. In comparison, DistilBERT lost 1.4\% test accuracy on the SST-2 task after losing 40\% of its parameters. This suggests that adding SideNets is a compute reduction method that can effectively complement knowledge distillation and model pruning.

As in the CIFAR10 case, we found that the SideNets were calibrated. However, we found that the pretrained transformers were also calibrated, duplicating the findings of  \cite{desai2020calibration}.

\subsection{Does the SideNet lower the MainNet's accuracy?}

It could be argued that the addition of the SideNet to the training task would lead to a decrease in the final accuracy of the MainNet, since the training procedure splits its attention between minimising the SideNet and the MainNet's loss. We find that this is not the case, and that adding the SideNet does not seem to have a negative effect on the MainNet's accuracy. Our findings are summarised in \autoref{summaryofmethods}. Anecdotally, we find that ensembling the SideNet and MainNet predictions provided a slight boost to final accuracy over just the MainNet's predictions. For a fuller discussion, see appendix F.

\begin{table}
  \caption{Test accuracies for final MainNet classifications when trained with and without a SideNet, on computer vision (CIFAR10) and natural language processing (SST-2) classification tasks. The values are averaged over 5 runs, and include standard deviations.}
  \label{summaryofmethods}
  \centering
  \begin{tabular}{ l  c c c c  }
    \toprule

     & ResNet18 & ResNet34 & ResNet50 & ResNet152   \\
    \midrule
    SideNet & $92.3 \pm .1$ & $93.2 \pm .1$ & $94.0 \pm .2$ & $94.3 \pm .1$\\
    No SideNet & $92.1 \pm .1$ & $92.9 \pm .2$ & $93.6 \pm .3$  & $93.9 \pm .2$  \\
    \bottomrule \smallskip
  \end{tabular}

\hspace{20em}

  \begin{tabular}{l  c c c}
     \toprule
     & DistilBERT & BERT-base & BERT-large \\
     \midrule
     SideNet & $88.8 \pm .1$ &  $90.6 \pm .2$ & $92.2 \pm 1.2$ \\
     No SideNet & $89.0 \pm .1$ & $90.6 \pm .3$ & $92.8 \pm 0.5$ \\
     \bottomrule
  \end{tabular}
\end{table}

\section{Conclusion and Future Work}
\label{concl}

In this work we propose attaching a SideNet, a small single hidden layer perceptron, onto the intermediate representations of a MainNet, a large pretrained network, and using the SideNet's confidence level to determine whether an input should be classified by the SideNet or passed back to the MainNet. SideNets are easy to implement, fine-tune, and deploy, and provide substantial compute savings at little cost to model accuracy, for both natural language processing and computer vision tasks. 

We also find that SideNets in the early layers of ResNets are calibrated, while the ResNets themselves are not, and that SideNets can significantly reduce the amount of compute used by DistilBERT at minimal cost to accuracy, despite DistilBERT already being a highly compressed model. Finally, increasing or decreasing the threshold $\theta$ for the model's confidence allows us to painlessly explore compute-accuracy space, by making continuous what was once discrete.

SideNets open several avenues for further study:

\begin{enumerate}
    \item SideNets perform well on classification tasks. Do they perform equally well on more complicated, higher dimensional tasks, such as image segmentation or machine translation?
    \item SideNets help reduce DistilBERT's total compute, with minimal loss in accuracy, even though DistilBERT is already a highly compressed model. What is the interplay between different forms of model compression, and to what extent can they be combined?
    \item SideNets are small and shallow networks. Does this make them more susceptible to being fooled by adversarial attacks \citep{goodfellow2014explaining}?
\end{enumerate}
 We hope to investigate these questions further in future work.

\section*{Broader Impact}

\subsection{Potential positive outcome}
We hope this paper demonstrates that conditional computation is a simple and effective tool for reducing the computational cost of neural networks, with minimal effect on model performance. We hope that our paper convinces machine learning researchers and engineers to adopt conditional computation methods in their own work--be it our simple SideNet method, or more powerful but involved methods like those described in our Related Work section (\autoref{relwork}). We think widespread adoption of conditional computation would be an effective tool in helping decrease deep learning's carbon footprint.

\subsection{Potential negative outcome}
We can imagine two ways in which SideNets could have a negative effect on society.
\begin{enumerate}
  \item SideNets are designed to draw quick conclusions from earlier-stage, less processed data. It is not clear whether or not the classifications they make would disproportionately leverage biases in the data and lead to unfair decisions.
  \item As discussed in \autoref{concl}, it might be easier to use adversarial examples to fool SideNets than to fool MainNets. If so, real-world systems that use them might be made more vulnerable by using conditional computing systems.  
\end{enumerate}
We hope to investigate both of these questions in future work, along with the open question posed in the conclusion.

\begin{ack}
We would like to thank Julien Raffaud, Patrick Germain, Tiffany Vlaar, Alexandre Matton, and Antreas Antoniou for helpful comments and conversations.
\end{ack}

\bibliographystyle{plainnat}
\bibliography{refs.bib}

\newpage

\begin{appendices}
\section{Softmax as confidence metric}

Suppose that a classifier outputs a vector $y = (y_1, \dots, y_{n})$. The softmax of $y$, which we call $\sigma(y)$, is an $n$-dimensional vector whose elements $\sigma(y)_i$ are computed in the following manner:

\[ \sigma (y)_i = \frac{e^{y_i}}{\sum\limits_{j=1}^{n} e^{y_j}} \]

\begin{figure}[ht]
  \centering
  \includegraphics[width=1\linewidth]{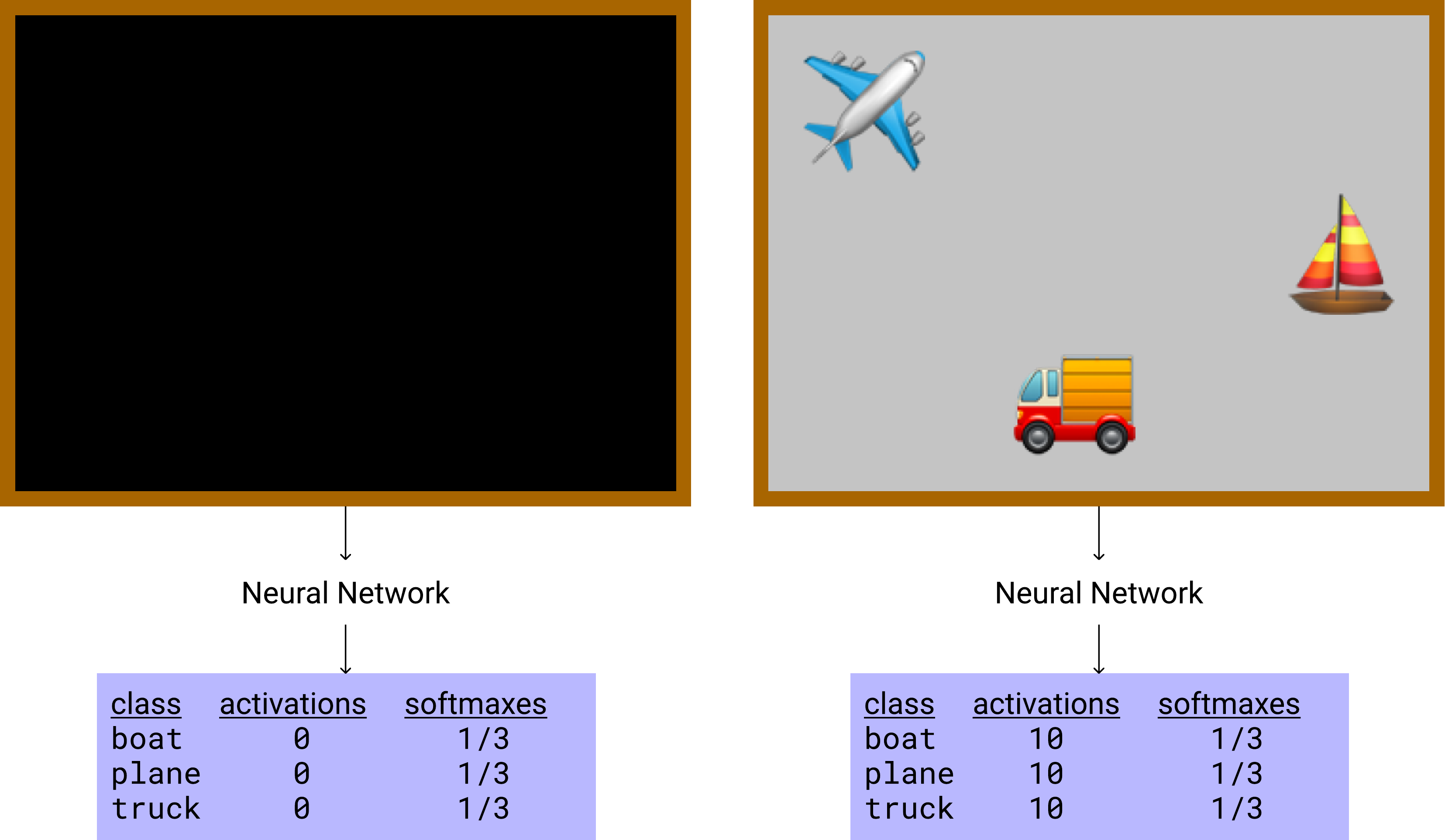}
  \caption{A neural network trained to classify images as containing boats, planes, or trucks, could give the same result for a picture containing simultaneously a boat, a plane, and a truck as it would for an entirely black picture.}
  \label{boatplanetruck}
\end{figure}

\autoref{boatplanetruck} illustrates the shortfalls of equating softmax outputs with model confidence: two images can be radically different and yet produce the same softmax output. Despite the theoretical shortcomings of softmax, we found nonetheless that it is a reasonable \textit{proxy} for model confidence, leading to good performance in our use-case, and producing calibrated results.

\section{Code snippets}

We stress that SideNets are remarkably easy to implement in practice. To define the models, we simply inherit from the original class, define the SideNet, and run it. The code to add a SideNet to ResNet50 and BERT-base is included below.

\begin{lstlisting}[language=Python, caption={SideNet on a ResNet50. The only code required to add a SideNet are lines 10-13, to define the SideNet, and line 20, to obtain the output of the SideNet.}]
from torchvision.models.resnet import ResNet, BasicBlock, Bottleneck
import torch.nn as nn
import torch.nn.functional as F

class MyResNet50(ResNet):
    
    def __init__(self, bottle=32):
        super(MyResNet50, self).__init__(Bottleneck, [3, 4, 6, 3])
        
        self.SideNet = nn.Sequential(nn.Linear(65536, bottle),
                                           nn.BatchNorm1d(bottle),
                                           nn.ReLU(),
                                           nn.Linear(bottle, 10))
        
        self.final_fc = nn.Linear(1000,10)
        
    def forward(self, x):
        out = F.relu(self.bn1(self.conv1(x)))
        out = self.layer1(out)
        SideNet_out = self.SideNet(out.view(-1, 65536))
        out = self.layer2(out.view(-1, 256, 16, 16))
        out = self.layer3(out)
        out = self.layer4(out)
        out = F.avg_pool2d(out, 2)
        out = out.view(out.size(0), -1)
        out = self.fc(out)
        out = self.final_fc(out)
        return out, SideNet_out
\end{lstlisting}

\begin{lstlisting}[language=Python, caption={SideNet on BERT-base. The code to add a SideNet are lines 22-26, to define the SideNet, and lines 30-32, to obtain the output of the SideNet.}]
import torch.nn as nn
import transformers.BertConfig as BC
import transformers.BertModel as BM
import BC.from_pretrained as fpt

class MyBert(nn.Module):
    
    def __init__(self, bottle=32):
        super(MyBert, self).__init__()
        
        uncased = `bert-base-uncased'
        config = fpt(uncased, output_hidden_states=True)

        self.model = BM.from_pretrained(uncased, config=config)
        
        self.fc = nn.Sequential(nn.Linear(768, bottle),
                                nn.ReLU(),
                                nn.BatchNorm1d(bottle),
                                nn.Linear(bottle, 1),
                                nn.Sigmoid())
        
        self.SideNet = nn.Sequential(nn.Linear(768, bottle),
                                     nn.ReLU(),
                                     nn.BatchNorm1d(bottle),
                                     nn.Linear(bottle, 1),
                                     nn.Sigmoid())
        
    def forward(self, input_id, attention_mask):
        out = self.model(input_id, attention_mask=attention_mask)
        intermediate = out[2][5].clone()
        # the <CLS> tags are at [:,0,:]
        SideNet_out = self.SideNet(intermediate[:,0,:])
        features = out[0][:,0,:]
        out = self.fc(features)
        return out, SideNet_out
        
\end{lstlisting}
        
\section{Changing number of hidden units in SideNet}

We find that increasing the number of hidden units in the SideNet does not affect results. If anything, it decreases accuracy (perhaps because it increases the number of parameters to train by more than an order of magnitude). Our findings are summarised in \autoref{bigvssmall}. 

\begin{table}[ht]
  \caption{Accuracy of DistilBERT's SideNet on SST-2, where the SideNet has an inner hidden layer of 768 neurons vs 32, with respect to the threshold $\theta$. We also vary the depth at which the SideNet is placed (we place it after the 1st, 2nd, 3rd, and 5th encoder block), and find that this has little effect. Results are averaged over 5 runs, with standard deviations.}
  \label{bigvssmall}
  \centering
  \begin{tabular}{ l  | r r r r r }
    \toprule

     & $\theta=20\%$ & $\theta=40\%$ & $\theta=60\%$ & $\theta=80\%$ & $\theta=90\%$  \\
    \midrule
    768 neurons (depth 1) &  $79.6 \pm 0.7$  & $82.8 \pm 0.7$  & $85.2 \pm 0.3$ & $86.9 \pm 0.5$ & $87.2 \pm 0.4$  \\
    32 neurons (depth 1) &  $81.4 \pm 0.7$  & $84.8 \pm 0.6$  & $86.9 \pm 0.5$ & $87.9 \pm 0.2$ & $88.5 \pm 0.5$\\
    \midrule
    768 neurons (depth 2) &  $79.7 \pm 0.4$ & $83.0 \pm 0.8$ & $85.0 \pm 0.5$ & $87.1 \pm 0.3$ & $87.4 \pm 0.4$ \\
    32 neurons (depth 2) &  $81.0 \pm 0.5$ & $84.7 \pm 0.1$ & $86.9 \pm 0.3$ & $88.1 \pm 0.4$ & $88.2 \pm 0.2$ \\
    \midrule
    768 neurons (depth 3) & $80.1 \pm 0.8$ & $82.8 \pm 0.6$ & $85.4 \pm 0.4$ & $86.5 \pm 0.6$ & $86.9 \pm 0.2$ \\
    32 neurons (depth 3) & $81.4 \pm 0.7$ & $84.7 \pm 0.5$ & $86.6 \pm 0.1$ & $88.2 \pm 0.2$ & $88.1 \pm 0.5$ \\
    \midrule
    768 neurons (depth 5) & $80.1 \pm 0.8$ & $83.1 \pm 0.6$ & $85.5 \pm 0.3$ & $86.8 \pm 0.5$ & $87.2 \pm 0.3$\\
    32 neurons (depth 5) & $81.5 \pm 0.6$ & $84.6 \pm 0.4$ & $86.5 \pm 0.4$ & $88.1 \pm 0.2$ & $88.2 \pm 0.1$ \\

    \bottomrule
  \end{tabular}
\end{table}

\section{Further experimental details}

In our experiments, we use PyTorch 1.5.0 \citep{paszke2017automatic}, torchvision 0.6.0 \citep{marcel2010torchvision}, and transformers 2.11.0 \citep{wolf2019transformers}. 

We use the same NVIDIA RTX 2070 GPU with 8GB of RAM for all our experiments. 

Our runtimes vary based on the models:

For the \textbf{text classification experiments}, it takes us 20 seconds per epoch to train DistilBERT, 40 seconds per epoch to train BERT-base, and 260 seconds per epoch to train BERT-large. DistilBERT is trained with batch size 32, BERT-base with batch size 16, and BERT-large with batch size 3 (due to memory constraints). 

For the \textbf{image classification experiments}, it takes us roughly 60 seconds for the smaller ResNets (18, 34, 50) and roughly 120 seconds for the larger ResNet152, all with the same batch size of 64. 

\section{More detailed diagrams}

\begin{figure}
  \centering
  \includegraphics[width=0.6\linewidth]{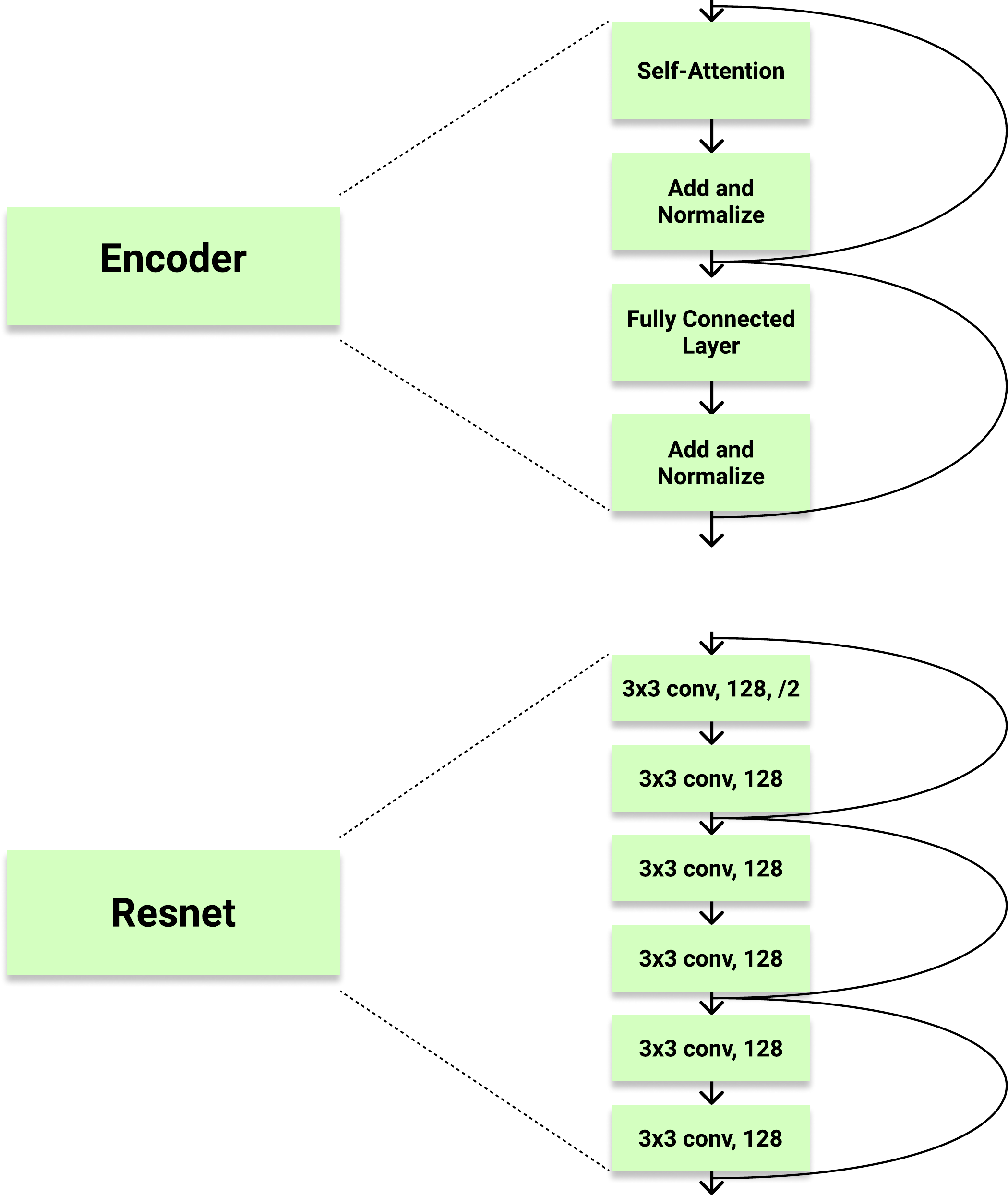}
  \caption{Details for what the (\textit{top}) Encoder and (\textit{bottom}) ResNet block contain in Figure 1.}
  \label{resnetnencoder}
\end{figure}

In Figure 1 we illustrate the general Transformer and ResNet architectures, without detailing the insides of the Encoder and Resnet blocks, due to the lack of space. We print them here in \autoref{resnetnencoder}, for completeness. 

\section{Ensembling Results}

We find that ensembling the predictions of the SideNet and the MainNet leads to a very slight boost in accuracy. Suppose our SideNet returns a vector $y_s = (y_{s1}, \dots, y_{s10})$ and our MainNet returns a vector $y_m = (y_{m1}, \dots, y_{m10})$. To classify the input, rather than return the argmax of  $y_m$, we take the argmax of $y_s+y_m = (y_{s1}+y_{m1}, \dots, y_{s10}+y_{m10})$.

Our results for the vision tasks are summarised in \autoref{ensemble}. The results weren't substantial enough to warrant our running experiments on text, but we include them here for the sake of completeness. 

\begin{table}
  \caption{CIFAR10 results for ensembling, averaged over 5 runs}
  \label{ensemble}
  \centering
  \begin{tabular}{ l  | r r r  }
    \toprule
    Model & MainNet Acc & SideNet Acc & Ensemble Acc  \\
    \midrule
    ResNet152 &  $94.44 \pm .11$  & $88.40 \pm .14$  & $94.52 \pm .19$ \\
    ResNet50 & $94.03 \pm .15$  & $89.14 \pm .18$  & $94.07 \pm .12$ \\
    ResNet34 & $93.21 \pm .09$  & $87.72 \pm .29$  & $93.20 \pm .07$ \\
    ResNet18 & $92.40 \pm .10$  & $86.79 \pm .49$  & $92.44 \pm .12$ \\
    
    \bottomrule
  \end{tabular}
\end{table}


\end{appendices}

\end{document}